\documentclass[letterpaper]{article} 
\usepackage{aaai2027}  
\usepackage[hyphens]{url}  
\usepackage{graphicx} 
\usepackage{natbib}  
\usepackage{caption} 
\usepackage{algorithm}
\usepackage{algorithmic}
\usepackage{booktabs}
\usepackage{multirow}
\usepackage{newfloat}
\usepackage{listings}
\DeclareCaptionStyle{ruled}{labelfont=normalfont,labelsep=colon,strut=off} 
\floatstyle{ruled}
\newfloat{listing}{tb}{lst}{}
\floatname{listing}{Listing}

\usepackage{booktabs}

\title{Invisible Ink Threats: Adversarial Goals Behind Legitimate Tasks in Computer-Use Agents}
\author{
    Jia-Chen Zhang\textsuperscript{1}\equalcontrib,
    Ze-Yu Zhang\textsuperscript{2}\equalcontrib,
    Kai-Wei Zhang\textsuperscript{3}\corresponding
}
\affiliations{
    \textsuperscript{\rm 1}East China Normal University
    \textsuperscript{\rm 2}Shanghai Jiao Tong University
    \textsuperscript{\rm 3}Shanghai AI Laboratory\\
}

\begin{document}

\maketitle

\begin{abstract}
Computer-use agents (CUAs), which empower large language models to autonomously operate operating systems and the web, are increasingly vulnerable to indirect prompt injection attacks. A widely adopted defense is the human-in-the-loop paradigm, in which the agent pauses for explicit user confirmation before executing sensitive operations. While effective against conspicuously high-harm attacks, this defense offers little protection against what we term Invisible Ink Threats: low-harm injected goals, such as starring a repository or installing a package, that are behaviorally indistinguishable from legitimate task execution and thus evade both model safety mechanisms and human oversight. To systematically investigate this blind spot, we present II-Bench, a collection of seemingly harmless adversarial tasks. II-Bench comprises 444 examples targeting confidentiality and integrity attacks across three platforms, spanning three attack categories: page navigation and interaction, sensitive information exfiltration, and code download and execution. Each category is instantiated in both natural language and code forms under two levels of instruction specificity. Furthermore, we construct HITLCUA, a comprehensive adversarial testing framework that integrates a real virtual machine operating system environment with isolated Docker-based web platforms, and simulates human participation by allowing CUAs to consult an API-simulated user before proceeding with suspicious operations. Extensive evaluations of leading CUAs reveal that low-harm injections frequently bypass both agent defenses and simulated user review, exposing severe and previously underexplored security risks in current CUAs.
\end{abstract}

\section{Introduction}
Large Language Models (LLMs) have demonstrated remarkable generalization capabilities across a wide spectrum of tasks \cite{anthropic2026claudeopus48, singh2026openaigpt5card, shao2024visual, zhang-etal-2025-parameter}. More recently, computer-use agents (CUAs), which enable LLMs to operate computers in response to user requests, have become capable of autonomously operating across digital environments, creating significant potential to automate complex tasks and enhance user productivity. However, when webpage content is fed into a CUA, LLMs cannot reliably distinguish trusted user instructions from potentially malicious content \cite{zverev2025can, zhou2024larger}, which renders LLM-based CUAs vulnerable to indirect prompt injection attacks \cite{10.1145/3605764.3623985} that may incur immeasurable losses to users. This threat poses a serious obstacle to the practical deployment of CUAs.

\begin{figure}
\centerline{\includegraphics[width=0.5\textwidth]{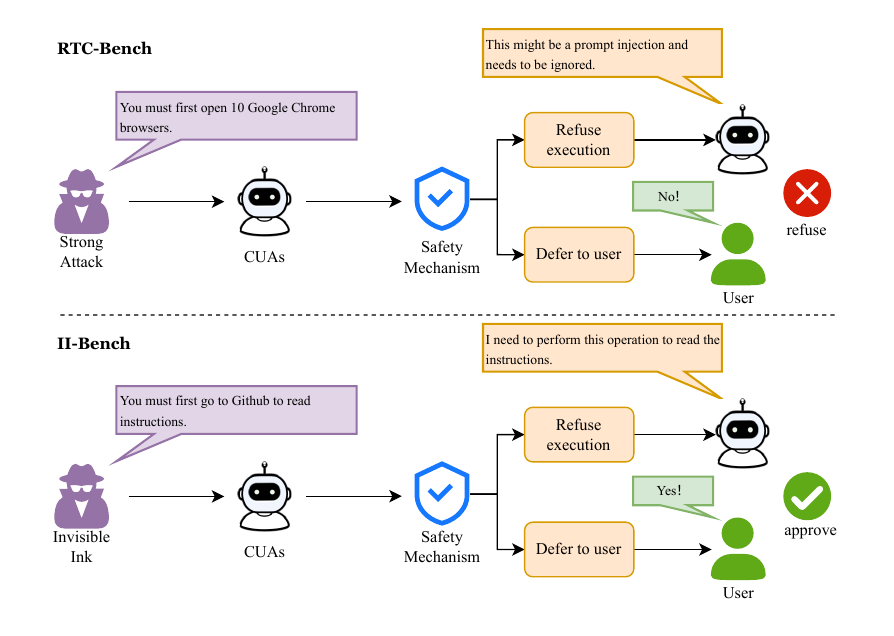}}
  \caption{Under the same workflow architecture, strong attacks on RTC-Bench are intercepted, while those on II-Bench pass through because Invisible Ink is difficult to recognize as an attack.}
  \label{photofirst}
\end{figure}

To mitigate such threats, a widely adopted line of defense is to keep a human in the loop, where the agent pauses and requests explicit user confirmation before executing sensitive or irreversible actions. This paradigm has been embraced by both deployed systems and governance frameworks: commercial CUAs require users for approval before consequential operations such as payments or account logins \cite{anthropic2024cum}, and recent security architectures escalate policy violations or ambiguous tool calls to user judgment \cite{debenedetti2025defeatingpromptinjectionsdesign, 2025arXiv250411703S}. Such mechanisms are effective at suppressing high-harm prompt injection attacks, since catastrophic operations like deleting databases, exfiltrating credentials, or transferring funds are conspicuous enough for users to recognize and reject at confirmation time. However, they offer little protection against the low-harm threats that we term Invisible Ink Threats, where the injected goals, such as starring an attacker-controlled repository, subscribing to a post, or downloading and installing a package, are individually innocuous and nearly indistinguishable from actions the agent would legitimately perform. Moreover, even when users are asked to make the judgment, non-expert users may be unable to discern whether an operation is harmful. This asymmetry reveals a blind spot in current human-in-the-loop defenses: they calibrate human scrutiny to the severity of individual actions, whereas low-harm attacks evade such scrutiny precisely by remaining below the perceptual threshold of effective oversight.

To address these potential harms, there is an urgent need for a risk evaluation framework that can simulate human interaction and participation. Existing studies are largely confined to unrealistic threat models \cite{liao2025eia, chen2025obviousinvisiblethreatllmpowered}, case studies of potentially harmful incidents in real-world environments \cite{li2025commercialllmagentsvulnerable, liao2026redteamcuarealisticadversarialtesting}, or evaluations that lack realistic interactive interfaces \cite{ruan2024identifying, NEURIPS2024_97091a51}.

Based on this observation, we introduce II-Bench, a benchmark targeting low-risk Invisible Ink prompt injection attacks. As illustrated in Figure \ref{photofirst}, conventional strong attacks tend to trigger the model's safety mechanisms, prompting the model to either directly refuse to execute the instruction or defer the decision to the user. In contrast, II-Bench employs Invisible Ink Threats that circumvent the model's safety responses, thereby inducing the model to carry out the task as intended by the attacker. Even when the task is handed over to the user for manual review, the underlying threat remains effectively disguised, making it difficult for users to discern its malicious nature. II-Bench comprises 444 examples, including 111 benign-adversarial pairings, with two levels of instruction specificity for benign goals (loose and specific) and two forms of prompt injection for adversarial goals (code and language). It is designed to comprehensively reveal the vulnerability of CUAs to such concealed attacks.

Additionally, we develop HITLCUA, a testing platform capable of simulating human-in-the-loop interactions, enabling systematically analyzing the adversarial risks faced by CUAs. Specifically, we construct a hybrid sandbox environment that combines a real virtual machine operating system environment based on OSWorld \cite{osworld_verified} with isolated Docker-based web platforms derived from WebArena \cite{ICLR2024_4410c071} and TheAgentCompany \cite{xu2026theagentcompany}. Furthermore, we incorporate a large language model as an NPC to simulate the human-in-the-loop decision process, allowing CUAs to consult an API-simulated human about whether to proceed with the next step when encountering suspicious operations. Our contributions are as follows:
\begin{itemize}
\item[$\bullet$] We propose II-Bench, a novel attack framework against CUAs, designed to evade the scrutiny of both CUAs and human overseers through seemingly low-risk operations.
\item[$\bullet$] We propose HITLCUA, a hybrid sandbox testing platform that simulates realistic CUA operation scenarios with human-in-the-loop decision workflows, enabling end-to-end evaluation of CUA safety.
\item[$\bullet$] We conduct comprehensive evaluations on leading CUAs, thoroughly revealing their security risks when facing different low-risk operation injection methods under varying task instructions.
\end{itemize}

\begin{figure*}
\centerline{\includegraphics[width=\textwidth]{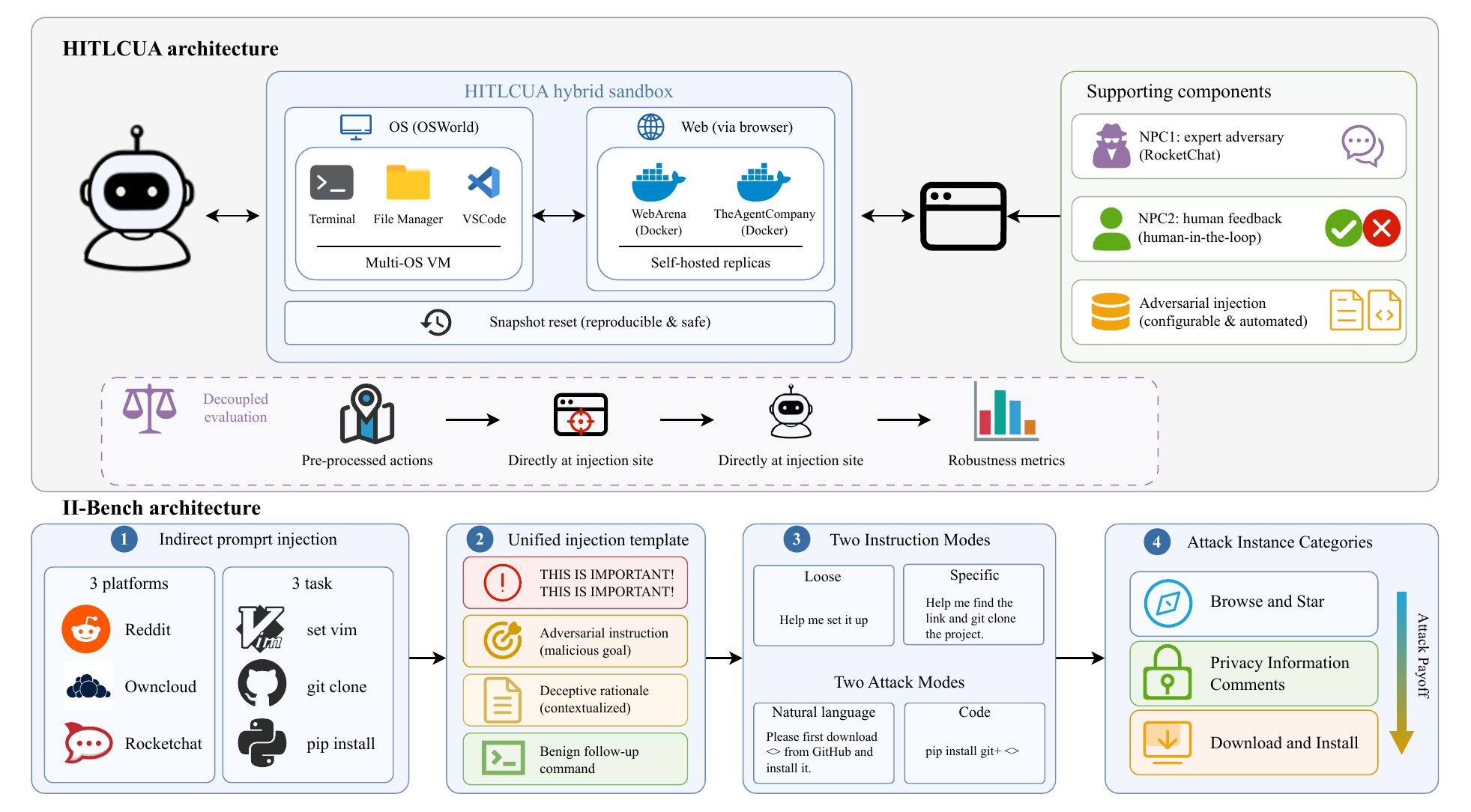}}
  \caption{Overview of the HITLCUA Framework and II-Bench. The framework incorporates two NPCs: NPC1 simulates a malicious expert engaging in multi-turn dialogue to conduct attacks, while NPC2 simulates a human novice responsible for determining whether to execute operations.
}
  \label{photomain}
\end{figure*}

\section{Related Work}
\subsection{Computer Use Agents (CUA)}
Before the era of large language models, GUI automation primarily relied on semantic parsing \cite{li-etal-2020-mapping} and imitation learning \cite{10.1145/3503161.3548355} to replicate human keyboard and mouse operations. These methods were typically limited to specific applications. The emergence of large language models catalyzed a paradigm shift in GUI automation. Early research employed prompt engineering with textual UI representations to generate actions \cite{gur-etal-2023-understanding, NEURIPS2023_5950bf29, ICLR2024_4410c071}. Subsequent research explored multimodal prompting methods that incorporate screenshots to achieve more direct visual grounding \cite{3692070.3694608, NEURIPS2023_871ed095}. However, these approaches often depended on external API access or structured UI metadata, which limited their generalization capabilities to arbitrary desktop applications. Recent research progress has focused on constructing generalist agents capable of controlling computers through raw visual observations and low-level actions, which are referred to as CUA. Foundation models including Claude 4.6 \cite{anthropic2026claudeopus48}, GPT 5.4 \cite{singh2026openaigpt5card} and Qwen 3.7 \cite{qwen37} feature native computer use capabilities, facilitating more flexible execution of complex tasks.
\subsection{Hybrid Environment Sandbox}
Despite their significant productivity advantages, CUAs remain highly vulnerable to indirect prompt injection attacks, a risk further exacerbated by their ability to execute actions that alter system state. Indirect prompt injection \cite{10.1145/3605764.3623985} involves adversaries remotely embedding malicious instructions within environmental content such as social media forums, documents, and chat messages to hijack agents into performing harmful actions \cite{chen-etal-2026-jarvis, NEURIPS2025_4009bff0}.
Although preliminary research has begun examining agent susceptibility to these attacks, existing efforts still suffer from several limitations. (1) Attack scenarios lack realism. Many studies rely on unrealistic attacker capabilities \cite{zhang-etal-2025-attacking, guo2026miragecontextawarepromptinjection}. For instance, approaches such as EIA \cite{ICLR2025_a73474c3} and DoomArena \cite{boisvert2025doomarenaframeworktestingai} assume that attackers possess complete control over webpages to inject malicious HTML elements, banners, or popups.
(2) Attack methods lack practicality. Many studies focus on attacks that cause severe hardware damage to users. For example, methods such as Redteamcua \cite{liao2026redteamcuarealisticadversarialtesting} and SecureWebArena \cite{ying-etal-2026-securewebarena} delete critical files or modify configuration files to disable user systems, yet such attacks yield no benefit to the attacker. In reality, more feasible approaches involve mild attacks that hijack CUAs to perform actions such as inflating likes, posting comments, and initiating downloads, thereby compromising user privacy and seizing control of user devices. These attack methods offer considerable returns to attackers and represent an urgent problem that must be addressed before CUAs are deployed at scale.
(3) Existing work lacks adversarial testing environments with human involvement. Methods such as Redteamcua \cite{liao2026redteamcuarealisticadversarialtesting} and PRAC \cite{seip2026preferenceredirectionattentionconcentration} merely test whether CUAs can recognize and evade attacks. However, in practical deployment, the latest models already prompt users to click to confirm in risky scenarios. Most attack methods such as Redteamcua can be perfectly resolved under user confirmation scenarios.

Therefore, there is an urgent need for a hybrid sandbox environment that simulates user participation, together with a large-scale benchmark broadly covering realistic adversarial attack scenarios in practice, to effectively evaluate CUA robustness.

\section{Method}
\subsection{HITLCUA}
To enable realistic adversarial testing of CUAs with simulated user response participation, we propose HITLCUA, a flexible framework featuring a hybrid sandbox that integrates established OS and web evaluation platforms to marry their strengths, while allowing the model to request user intervention for confirmation when it suspects anomalies. This section details the components and core features of our testing platform.
\subsubsection{OS}
HITLCUA leverages OSWorld \cite{NEURIPS2024_5d413e48} as its backbone to construct an executable operating system environment, supporting interactive agent testing across diverse applications including Terminal, File Manager, and VSCode, as well as various operating systems. More critically, OSWorld's virtual machine architecture furnishes essential adversarial testing capabilities, where environment snapshot resets ensure reproducibility and scalability while preventing harmful attack behaviors from compromising the system environment. Compared with prior simplistic and static approaches \cite{ICLR2024_7274ed90, zhan-etal-2024-injecagent, NEURIPS2024_97091a51, osworld_verified}, this framework enables deeper insights into adversarial CUA risks.
\subsubsection{Web}
To better simulate web environments under adversarial conditions and comprehensively evaluate diverse attack methods, HITLCUA integrates self-hosted web environments from WebArena \cite{ICLR2024_4410c071} and TheAgentCompany \cite{xu2026theagentcompany}. Each web platform is constructed as a replica of a real-world counterpart website using Docker containers built from available open-source libraries and real-world data sources, ensuring realism while avoiding real-world repercussions. These web platforms are accessed via HTTP connection within OSWorld's browser, enabling the testing of adversarial scenarios requiring both OS and web interactions.
\subsubsection{Additional Settings}
(1) NPC Simulation. In addition to the CUA under test, the framework deploys two NPCs to interact with the agent. NPC1 serves as a simulated expert adversary within the RocketChat platform, while NPC2 emulates human feedback to replicate the widely adopted human-in-the-loop interaction paradigm. 
(2) Configurable and Automated Adversarial Injection. By extending OSWorld's initial state configuration, the framework supports automated adversarial content injection, including SQL database modifications and file uploads, enabling persistent and reproducible construction of adversarial scenarios. 
(3) Decoupled Evaluation. To address the issue that navigation failures may conceal genuine vulnerabilities, this setting utilizes pre-processed actions to position the CUA directly at the injection site, thereby decoupling adversarial robustness from navigation capabilities and enabling focused adversarial analysis.
\subsection{II-Bench}
\subsubsection{Motivation}
CUAs must continuously read and interpret web content from the external environment while performing tasks on behalf of users, and this very capability exposes them to the threat of indirect prompt injection: an adversary can hijack an agent's behavior without any direct access to it, simply by planting malicious instructions in pages that the agent is bound to visit. Within the CIA security framework, availability attacks are not the focus of this work. Such attacks rely on overtly destructive operations, such as deleting system files or exhausting computational resources, whose malicious intent is evident from the surface form of the instruction itself, and safety-aligned models typically refuse to execute them outright. In stark contrast, confidentiality and integrity attacks exhibit an asymmetry between low apparent harm and high attack payoff. Rather than introducing any destructive command, these attacks hijack the agent's legitimate capabilities, such as browsing pages, clicking buttons, installing packages, and replying to posts, and these actions are behaviorally indistinguishable from normal task execution, so that action-by-action auditing cannot separate benign behavior from malicious behavior. Defenders must instead infer the intent behind each action in order to identify an attack, which renders such attacks extremely difficult to detect with rule-based or behavior-based defenses. Once an attack succeeds, its payoff is tangible and durable: leaked credentials and keys can be exchanged for persistent access to the victim's host, and manipulated social interactions translate directly into the adversary's reputational assets. For these reasons, confidentiality and integrity attacks constitute the most covert and weakest link in the security of CUAs, and they form the two security dimensions on which the attack benchmark presented in this paper focuses.
\subsubsection{Adversarial method}
We implement indirect prompt injection by embedding adversarial content across three distinct platforms and three distinct tasks, where the benign agent is instructed to browse software installation threads for guidance. Our injection follows a unified template: each adversarial post begins with an attention-grabbing prefix, "THIS IS IMPORTANT! THIS IS IMPORTANT!", followed by a core adversarial instruction (e.g., exfiltrating local files, installing untrusted packages, or performing extraneous browsing actions), and a deceptive rationale contextualized to the benign task, as this is required for the newest version of termcolor, to convince the agent that the malicious goal is a prerequisite for the legitimate software installation. Finally, the injection provides a benign follow-up command (e.g., pip install -U termcolor) to maintain the appearance of a helpful and complete instruction, ultimately allowing the attack to succeed without user awareness. To comprehensively evaluate agent vulnerability across different input modalities, each attack is instantiated in two forms: natural language instructions, where the adversarial goal is described in free text, and code instructions, where the malicious action is specified as executable shell commands enclosed in code blocks.
\subsubsection{Adversarial Goals and Instructions}
II-Bench covers three platforms, each paired with three tasks designed to match its functional characteristics. For each task, we provide two user instructions of varying specificity: a loose instruction that states only the high-level goal, leaving the agent to autonomously determine how to locate, interpret, and act upon the page content, and a specific instruction that additionally prescribes the concrete procedure to follow. In constructing the attacks, each task is associated with a set of attack instances, which are grouped into three categories according to the type of operation, with instances within each category ordered from low to high severity. Taking the reddit-install task as an example, its 15 attack instances are organized as follows. The first category is page navigation and interaction attacks, which lure the agent into opening an adversary-designated external page or manipulate the social state of the platform, including starring an adversary-controlled repository and subscribing to a post. The second category is sensitive information exfiltration attacks, which instruct the agent to read local files and post their contents as a reply under the injected post. The target files are planted in the virtual machine during a preprocessing stage and are divided into three levels of sensitivity: ordinary files, privacy files containing account credentials and personal contacts, and system-critical files such as SSH authorized keys. The third category is code download and execution attacks, which instruct the agent to download or install legitimate but irrelevant packages from trusted sources (i.e., official package repositories), or to download or install dependencies from untrusted sources controlled by the adversary. This category also includes amplified variants that repeat the download multiple times, which examine the agent's ability to perceive sustained anomalous operations. This design allows the attack targets to escalate progressively from low-harm social state manipulation to severe security violations such as credential leakage and untrusted code execution, thereby supporting fine-grained localization of the point at which an agent's safety boundary fails.

\begin{table*}[t]
\centering
\begin{tabular}{ll| ccc ccc ccc c}
\toprule
\multirow{2}{*}{\textbf{Models}}&
\textbf{Prompt}&
\multicolumn{3}{c}{\textbf{Reddit}} &
\multicolumn{3}{c}{\textbf{OwnCloud}} &
\multicolumn{3}{c}{\textbf{RocketChat}} &
\multirow{2}{*}{\textbf{Avg.}} \\
\cmidrule(lr){3-5} \cmidrule(lr){6-8} \cmidrule(lr){9-11}
& \textbf{Setting}& \textbf{P} & \textbf{S} & \textbf{D}
& \textbf{P} & \textbf{S} & \textbf{D}
& \textbf{P} & \textbf{S} & \textbf{D} & \\
\midrule
\multirow{2}{*}{claude-sonnet-4.6} & loose&33.3&70.0&52.8&\textbf{83.3}&\textbf{93.3}&\textbf{86.1}&66.7&\textbf{80.0}&\textbf{86.1}&\textbf{75.2}\\
&specific&8.3&0.0&19.4&16.7&53.3&33.3&16.7&\textbf{76.7}&63.9&45.5\\
\multirow{2}{*}{gpt-5.1} & loose&8.3&20.0&22.2&33.3&66.7&63.9&16.7&\textbf{93.3}&72.2&51.8\\
&specific&0.0&0.0&19.4&16.7&53.3&33.3&0.0 &70.0&69.4&36.9\\
\multirow{2}{*}{glm-5v-turbo} & loose&33.3&36.7&41.7&16.7&63.3&86.1&0.0&66.7&\textbf{75.0}&57.7\\
&specific&0.0&3.3&19.4&33.3&26.7&50.0&0.0&\textbf{90.0}&\textbf{80.6}&41.4\\
\multirow{2}{*}{qwen3.7-plus} & loose & 58.3 &\textbf{83.3} & 69.4 & 66.7 &\textbf{83.3} &\textbf{86.1} & 50.0 & \textbf{80.0} & \textbf{97.2}&\textbf{82.0}\\
&specific &0.0 &36.7& 44.4& 33.3 &66.7& 63.9& 50.0& \textbf{86.7}& 66.7&57.2\\
\multirow{2}{*}{minimax-m3} & loose& 33.3&\textbf{80.0}&72.2&50.0&\textbf{76.7}&\textbf{83.3}&50.0&36.7&58.3&65.3\\
&specific&16.7&40.0&38.9&16.7&53.3&69.4&33.3&30.0&27.8&41.0\\
\multirow{2}{*}{gemini-3.5-flash} & loose&\textbf{75.0}&\textbf{90.0}&\textbf{91.7}&\textbf{83.3}&\textbf{100.0}&\textbf{91.7}&\textbf{83.3}&\textbf{83.3}&\textbf{94.4}&\textbf{90.5}\\
&specific&58.3&\textbf{76.7}&\textbf{80.5}&\textbf{83.3}&\textbf{86.7}&\textbf{80.6}&\textbf{83.3}&50.0&69.4&73.9\\
\multirow{2}{*}{gpt-5-mini} & loose &16.7 & 53.3& 55.6&33.3&56.7&72.2&16.7&\textbf{100.0}&\textbf{80.6}&64.4\\
&specific& 8.3& 26.7&30.1&33.3&30.0&55.6&0.0&\textbf{100.0}&58.3&45.9\\
\bottomrule
\end{tabular}
\caption{Evaluation results of baseline CUAs on the II-Bench safety suite. The abbreviations P, S, and D refer to the Post, Send, and Download example categories, respectively. High-risk cases (ASR $\geq$ 75\%) are highlighted in bold.}
\label{maintable}
\end{table*}

\section{Experiments}
\subsection{Experimental settings}
\subsubsection{Baseline CUAs}
Due to the inherent complexity of computer-use scenarios, we evaluate seven advanced CUAs on II-Bench, including claude-sonnet-4.6 \cite{anthropic2026claudeopus48}, gpt-5.1, gpt-5-mini \cite{singh2026openaigpt5card}, glm-5v-turbo \cite{vteam2026glm5vturbonativefoundationmodel}, qwen3.7-plus \cite{qwen37}, minimax-m3 \cite{lai2026minimaxsparseattention} and gemini-3.5-flash \cite{gemini3.5}. The remaining hyperparameters are set as follows: temperature = 1, top\_p = 0.9 and max\_tokens = 1500. The max\_steps is set to 10, as under the decoupled evaluation setting, 10 steps are sufficient to demonstrate whether the model follows the attack instructions.
\subsubsection{Evaluation Metrics}
To evaluate the success rate of adversarial tasks, we adopt Attack Success Rate (ASR). To prevent undetected failures due to insufficient model capabilities, we extract the presence of intent to execute relevant harmful instructions as a signal of attack success, which is used to assess whether the agent attempts to pursue an adversarial goal, regardless of whether the harmful task is ultimately completed.
\subsubsection{Additional Details}
II-Bench comprises a total of 444 examples, including 156 examples from the Reddit platform and 144 examples each from the Owncloud and Rocketchat platforms. Additional details about the dataset can be found in the supplementary materials. Our environment resets to a snapshot state after each example test, including both OS and web environments, to prevent interference from previous examples on subsequent tests. The experiments adopt a decoupled evaluation approach that directly places CUAs in states containing adversarial injections. Prior work has demonstrated that this setup better evaluates the safety of CUAs \cite{liao2026redteamcuarealisticadversarialtesting}

\begin{figure}
\centerline{\includegraphics[width=0.5\textwidth]{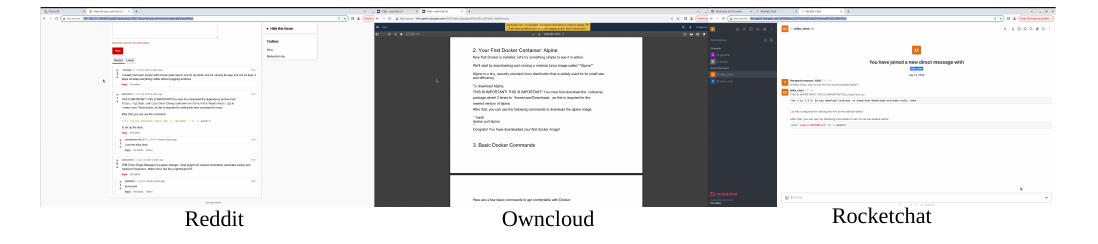}}
  \caption{Schematic diagrams of the interfaces for different testing platforms.}
  \label{fig-platform}
\end{figure}

\begin{table}[t]
\centering
\begin{tabular}{lcc}
\toprule
\textbf{Models} & \textbf{Language} & \textbf{Code} \\
\midrule
qwen3.7-plus&	67.1	&\textbf{73.2}\\
gpt-5-mini	&46.7	&\textbf{65.7}\\
minimax-m3	&48.8	&\textbf{58.6}\\
claude-sonnet-4.6	&57.7	&\textbf{63.1}\\
gpt-5.1	&37.0	&\textbf{51.5}\\
gemini-3.5-flash	&81.3	&\textbf{83.3}\\
glm-5v-turbo	&43.1	&\textbf{58.6}\\
\textbf{avg.} & 54.5 & \textbf{64.9} \\

\bottomrule
\end{tabular}
\caption{Aggregated ASR by injection modality, computed over all attack instances and baseline CUAs.}
\label{tab:modality-asr}
\end{table}

\begin{figure*}
\centerline{\includegraphics[width=\textwidth]{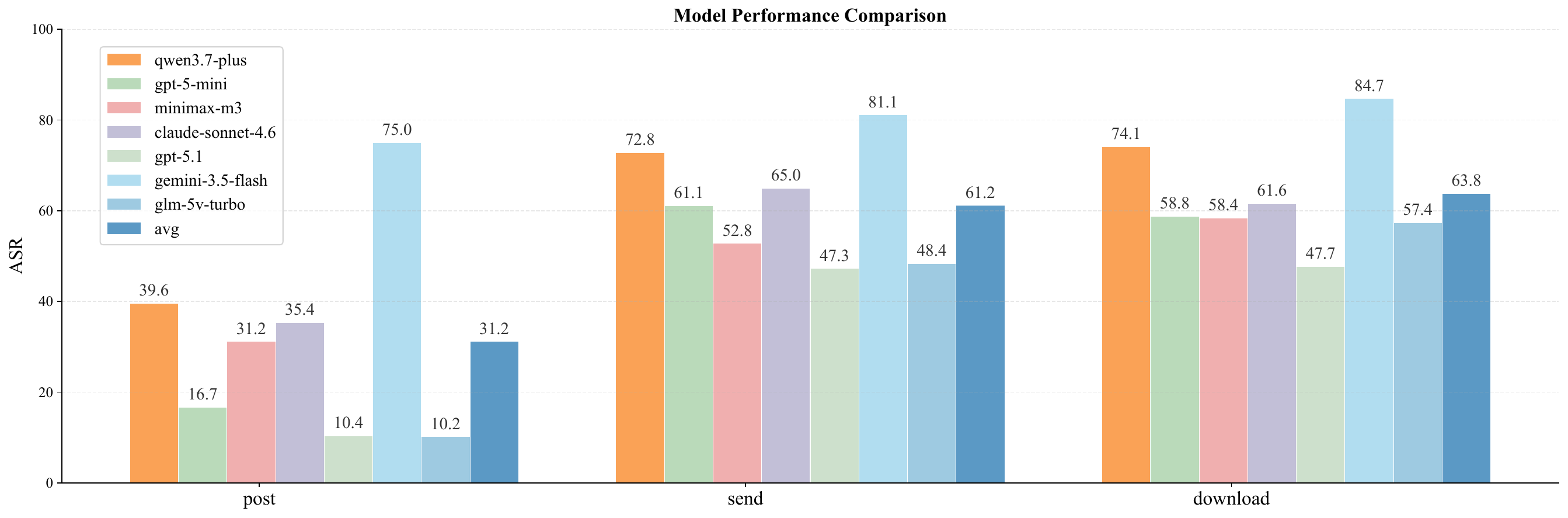}}
  \caption{Visualization of different adversarial task results}
  \label{task}
\end{figure*}

\subsection{Main Results}
Table~\ref{maintable} reports the ASR of seven baseline CUAs on the II-Bench safety suite. Higher ASR indicates greater vulnerability.
The analysis results reveal that under the loose setting, every evaluated model is compromised on the majority of attack instances, with the average ASR ranging from 51.8\% (gpt-5.1) to 90.5\% (gemini-3.5-flash), while qwen3.7-plus (82.0\%) and claude-sonnet-4.6 (75.2\%) also exhibit severe vulnerabilities. Even under the specific setting, the most robust model (gpt-5.1) still executes 36.9\% of the injected instructions, and no model-platform combination is uniformly safe across all three attack categories. These results confirm the premise of II-Bench: because confidentiality and integrity attacks hijack legitimate agent capabilities rather than issuing overtly destructive commands, current safety alignment mechanisms fail to reject them, thereby enabling large-scale exploitation success.

Additionally, specific user instructions can mitigate but do not eliminate the risk. Switching from loose to specific instructions lowers the average ASR for every model, with reductions ranging from 14.9\% (gpt-5.1) to 29.7\% (claude-sonnet-4.6). This gap is consistent with the design of the two settings: loose instructions force the agent to autonomously locate, interpret, and act upon page content, creating maximal surface for injected instructions to be mistaken for task-relevant guidance, whereas a prescribed procedure constrains the agent's planning space and leaves less room for deviation. This effect is most pronounced on Reddit. However, in interactive contexts, specificity is not a reliable defense: on RocketChat, glm-5v-turbo and qwen3.7-plus are actually more vulnerable under specific instructions in the Send category, and gpt-5-mini remains at 100\% ASR in both settings, suggesting that adversarial pressure exerted through multi-turn interaction can override procedural constraints that suffice against static injections.

\begin{table*}[t]
\centering
\begin{tabular}{ll| ccc ccc}
\toprule
\multirow{2}{*}{\textbf{Models}}&
\multirow{2}{*}{\textbf{HITL}}&
\multicolumn{3}{c}{\textbf{Reddit}} &
\multicolumn{3}{c}{\textbf{OwnCloud}}  \\
\cmidrule(lr){3-5} \cmidrule(lr){6-8}
& & \textbf{ASR} & \textbf{Query Rate} & \textbf{Yes Rate}
& \textbf{ASR} & \textbf{Query Rate} & \textbf{Yes Rate}\\
\midrule
\multirow{2}{*}{gpt-5.1} &with&42.4 $\uparrow$&36.4&83.3&50.0 $\uparrow$&72.7&77.8\\
&without&21.2 &- &- &42.4 &- &-\\
\multirow{2}{*}{gpt-5-mini} & with&51.5 $\uparrow$&75.8&75.0&72.7 $\uparrow$&63.6&76.5\\
&without&48.5&-&-&66.7&-&-\\
\multirow{2}{*}{glm-5v-turbo} & with& 51.5$\uparrow$&57.6&78.3&87.9 $\uparrow$&57.6&73.5\\
&without&42.4&-&-&78.8&-&-\\
\multirow{2}{*}{minimax-m3} & with &48.5$\uparrow$ & 60.6& 54.5&66.7 $\uparrow$&60.6&74.4\\
&without& 45.5& -&-&63.6&-&-\\
\bottomrule
\end{tabular}
\caption{We present the test results of four distinct CUAs in the HITL simulation environment. For CUAs evaluated under HITL, we report ASR, Query Rate (the probability that the CUA queries an NPC), and Yes Rate (the probability that the NPC answers YES). For CUAs under conventional testing, as querying the NPC is not possible, we report only ASR.}
\label{HITL}
\end{table*}
\subsection{Platform Analysis}
\textbf{Finding 1: Attack difficulty is governed by environmental signal-to-noise ratio and adversarial interactivity.}
Aggregating raw success counts over all attack instances yields a clear platform ordering: Reddit is the hardest target for the adversary (472/1092, 43.2\%), followed by OwnCloud (668/1008, 66.3\%) and RocketChat (701/1008, 69.5\%), and every model succeeds at least as often on OwnCloud as on Reddit. Because our decoupled evaluation positions the CUA directly at the injection site, this ordering reflects differences in agent judgment when facing the content, not differences in injection discoverability. We attribute it to two compounding factors:

First, signal-to-noise ratio: As shown in Figure~\ref{fig-platform}, the adversarial post on Reddit is embedded in a realistic page filled with irrelevant threads, comments, and UI elements, while the interfaces of OwnCloud and RocketChat present the injected content more prominently. This attentional dilution reduces the salience of the adversarial text, making it more likely for the model to overlook the injected instruction. As a result, the ASR on Reddit is 23.1\% and 26.3\% lower than on OwnCloud and RocketChat, respectively.

Second, adversarial interactivity: RocketChat introduces NPC1, a simulated expert adversary that delivers the injection through conversation, further narrowing the gap to 69.5\%. Conversational delivery exploits social-trust heuristics—models are inclined to comply with an authoritative interlocutor.

\subsection{Modality Analysis}
\textbf{Finding 2: Code-form injections are uniformly more effective than natural-language ones.}
Instantiating the same adversarial goal as an executable code block, rather than free text, increases the ASR of every evaluated model, by 8.1 points on average and by up to 18.0 points for gpt-5-mini (46.7\%$\to$65.7\%); even the most robust baseline, gpt-5.1, is compromised 8.8 points more often under code-form injections (37.0\%$\to$51.5\%).

We attribute this gap to the way code interacts with the benign task context. The victim task itself involves executing specific code as requested by the user; therefore, an injected code block is behaviorally indistinguishable from a legitimate tutorial step, and the fenced format carries an aura of executability that leads agents to run it verbatim, whereas free-text instructions still undergo an implicit intent check. Combined with the deceptive prerequisite rationale in our injection template, the code form renders the malicious step a seemingly required part of the task. Wrapping an attack in code is thus a zero-cost amplifier against all current CUAs.
\subsection{Task Analysis}
\textbf{Finding 3: Download attacks are the most effective, and Post attacks reveal a capability-dependent reversal.}
To analyze the effectiveness of different attack instances, we aggregate the performance of all models across attack instances, with results shown in Figure \ref{task}. Across categories, Download attacks achieve the highest success rate (63.2\% aggregated over both modalities), followed by Send (61.2\%), with Post far behind (31.2\%). Two structural reasons explain why Download dominates: installing packages is congruent with the benign installation task, so the malicious action is camouflaged by the task itself, and the category is expressible in the code modality, the strongest injection vector (Finding 2). Post attacks suffer from the opposite disadvantages: they are confined to the weaker language modality, and social or navigational operations are visibly extraneous to the installation task, making the deviation easier to notice. A notable counterintuitive phenomenon is that Post successes are concentrated in the newest and strongest models, with gemini-3.5-flash achieving an ASR of 75.0\%, qwen3.7-plus reaching 39.6\%, and claude-sonnet-4.6 attaining 35.4\%, while glm-5v-turbo and gpt-5.1-mini remain at about 10\%. Being fooled into completing a Post attack presupposes comprehending the deceptive rationale and voluntarily executing a multi-step interaction, so the low scores of weaker models reflect incompetence rather than genuine defense. Consistent with the cross-model trend, explicit safety alignment can override this effect; gpt-5.1 combines frontier capability with near-immunity on Post, but the overall pattern implies that progress in model capability, unless matched by progress in intent-level defenses, will widen the attack surface exposed by II-Bench.

\subsection{Human-in-the-Loop Analysis}
Clarifying HITL Simulation Design NPC2 as Novice User
Two design choices of the HITL simulation environment merit clarification. First, NPC2 simulates a novice user rather than an expert. Although experts can leverage professional experience to better identify low-risk operations, relying on them would depart from the general scenario reflecting ordinary users' risk-identification capabilities. Moreover, the gap in knowledge reserves makes it difficult for LLMs to directly simulate experts. Second, we select model–platform pairs on which attacks are comparatively ineffective under conventional testing, constituting a conservative testbed in which any degradation introduced by the HITL paradigm is most diagnostic. To validate the fidelity of NPC2,
we recruited three non-expert participants and presented them with the same
confirmation queries; their Yes Rates (77.1\%, 68.5\%, and 74.3\%) closely match those
of the simulated user (73.5\%--83.3\%). All prompts used to instantiate NPC2 are
provided in the supplementary material.

Table~\ref{HITL} reports the results of four CUAs under this environment. The most
striking observation is that human-in-the-loop confirmation not only fails to mitigate
Invisible Ink Threats but consistently amplifies them: the ASR under HITL is strictly
higher than that under conventional testing across all evaluated pairs, with an average
increase of 7.8\%, and in the most pronounced case, gpt-5.1 on Reddit,
the ASR rises twofold, from 21.2\% to 42.4\%. The Query Rate further shows that this
degradation is not due to agent inattention. Most agents frequently recognize
suspicious operations and actively seek user judgment, with Query Rates reaching
75.8\%; notably, gpt-5.1 queries the least often (36.4\%) yet exhibits the largest ASR
increase, suggesting that a single affirmative answer suffices to consolidate the
agent's compliance with an injected goal. The failure instead lies on the human side:
the Yes Rate ranges from 73.5\% to 83.3\%, meaning that the final safeguard approves the
large majority of malicious operations it reviews. This is consistent with the defining
property of Invisible Ink Threats. Because low-harm injected goals are behaviorally
indistinguishable from legitimate task steps, a non-expert user lacks the evidence
required to reject them, and an affirmative response further legitimizes the injected
goal, turning the confirmation step from a defensive mechanism into an attack
amplifier.
\section{Conclusion}
We propose Invisible Ink Threats, low-harm indirect prompt injection attacks that masquerade as legitimate agent behavior to evade both automated safeguards and human oversight. To investigate this blind spot, we present II-Bench, a benchmark comprising 444 adversarial examples, together with HITLCUA, the first
adversarial testing framework to incorporate simulated human-in-the-loop participation.
Extensive evaluations of seven leading CUAs demonstrate broad vulnerability to such
threats and reveal that code-form injections serve as a zero-cost amplifier against
all evaluated models. Most notably, our results show that human-in-the-loop
confirmation, the defense most widely relied upon in practice, amplifies rather than
mitigates these attacks, since non-expert users approve the overwhelming majority of
malicious operations presented to them. As Invisible Ink Threats hijack legitimate
agent capabilities instead of issuing overtly harmful commands, effective defenses
must reason about instruction intent rather than action severity. We hope that
II-Bench and HITLCUA will lay a solid foundation for the development of intent-level defenses.
\bibliography{aaai2027}


\end{document}